\documentclass[sigconf, nonacm]{acmart}

\usepackage{enumitem}

\AtBeginDocument{
  }

\setcopyright{acmlicensed}
\copyrightyear{2018}
\acmYear{2018}
\acmDOI{XXXXXXX.XXXXXXX}

\acmConference[Conference acronym 'XX]{Make sure to enter the correct
  conference title from your rights confirmation email}{June 03--05,
  2018}{Woodstock, NY}

\acmISBN{978-1-4503-XXXX-X/2018/06}

\usepackage{tcolorbox}
\tcbuselibrary{listings}
\usepackage{booktabs}
\usepackage{multirow}
\usepackage[table,xcdraw]{xcolor}

\begin{document}

\title{MACReD: A Multi-Agent Collaborative Reasoning \\ Framework for Reaction Diagram Parsing}

\author{Chuang Tang}
\email{tangchuang@mail.ustc.edu.cn}
\affiliation{
  \institution{University of Science and Technology of China}
  \city{Hefei}
  \country{China}
}

\author{Chenhao Lin}
\email{mojizooo@mail.ustc.edu.cn}
\affiliation{
  \institution{University of Science and Technology of China}
  \city{Hefei}
  \country{China}
}
\author{Yin Xu}
\author{Hao Wang}
\email{{yinxu,wanghao3}@ustc.edu.cn}
\affiliation{
  \institution{University of Science and Technology of China}
  \city{Hefei}
  \country{China}
}

\author{Jinrui Zhou}
\email{zzkevin@mail.ustc.edu.cn}
\affiliation{
  \institution{University of Science and Technology of China}
  \city{Hefei}
  \country{China}
}

\author{Xin Li}
\email{leexin@ustc.edu.cn}
\affiliation{
  \institution{University of Science and Technology of China \& iFLYTEK Co., Ltd.}
  \city{Hefei}
  \country{China}
}

\author{Mingjun Xiao}
\author{Enhong Chen}
\email{{xiaomj,cheneh}@ustc.edu.cn}
\affiliation{
  \institution{University of Science and Technology of China}
  \city{Hefei}
  \country{China}
}

\renewcommand{\shortauthors}{Tang et al.}

\begin{abstract}

  Parsing chemical reaction diagrams from scientific literature is challenging due to heterogeneous layouts, intertwined visual elements, and difficulty of integrating recognition and reasoning. 
  Existing Vision Language Models advance multimodal understanding but fail on complex diagrams, struggling to maintain spatial coherence and failing to integrate multidimensional information during reasoning.
  To address these issues, we propose \textbf{MACReD}, a hierarchical multi-agent framework that coordinates specialized agents for molecular perception, arrow understanding, text extraction, and reaction reconstruction a unified VLM-guided architecture. Planning and Perception Layers use flexible, fine-grained detection to handle visual complexity, while the Reasoning Layer uses a Multigraph Fusion mechanism to integrate heterogeneous cues and enforce chemically consistent global reasoning. Experiments on RxnScribe benchmark show \textbf{MACReD} achieves state-of-the-art performance, with F1 scores of 75.2\% and 84.6\% under hard and soft match criteria, outperforming baseline RxnScribe (69.1\% / 80.0\%) and demonstrating robustness across diverse diagram layouts, including multi-step and tree-structured reactions.
  Our code\footnote{\url{https://github.com/TC9905/MACReD}} is publicly available.

\end{abstract}


\maketitle

\vspace{-0.1in}
\section{Introduction}
\label{sec:introduction}

\begin{figure}
\centering
\includegraphics[width=0.9\linewidth]{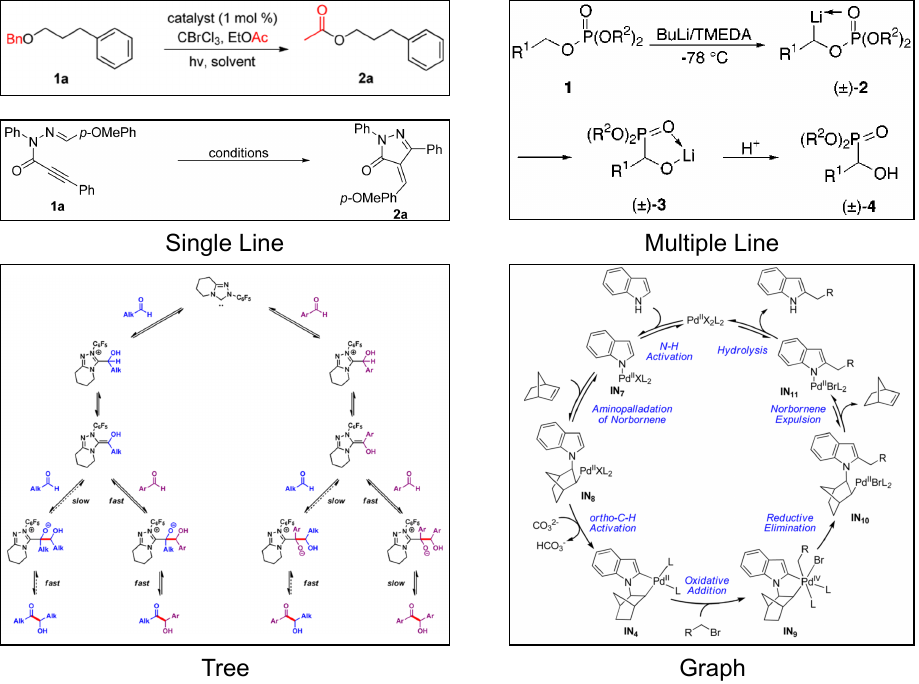}
\vspace{-0.1in}
\caption{Examples of reaction diagrams in chemistry literature, from simple single-step to complex pathways.}
\label{fig:diagrams_type}
\vspace{-0.2in}
\end{figure}

The rapid evolution of AI for Science (AI4Sci) has catalyzed a transformative shift in organic chemistry, with machine learning now driving breakthroughs in retrosynthesis, reaction prediction, and condition optimization \cite{ding2025survey,back2024accelerated}. However, the efficacy of these models is fundamentally predicated on the availability of high-quality, machine-readable datasets. While scientific literature serves as a vast repository of chemical knowledge, this information remains largely ``locked'' within sophisticated multimodal graphics. Among the various forms of chemical graphics, reaction diagrams constitute one of the most information-dense and widely used representations. Unlike structured data, these diagrams interweave molecular structures, reaction arrows, reagents, and intermediate states. As illustrated in Figure ~\ref{fig:diagrams_type}, diagrams have evolved beyond simple single-step transformations to encompass complex, multi-step, and branched synthetic pathways. More importantly, the visual layout of a reaction diagram does not directly reflect its underlying reaction logic, e.g., spatial proximity does not necessarily indicate chemical relevance, arrows may convey diverse reaction semantics, and intermediate states are often implied rather than explicitly annotated. While this representation is intuitive for human experts, automated parsing requires machines to transcend basic visual recognition to infer latent reaction structures and resolve ambiguities under global chemical consistency constraints. Therefore, \textit{there is an urgent need for a robust, chemically-aware parsing framework capable of decoding these sophisticated visual hierarchies.}

Early studies mainly leveraged Optical Chemical Structure Recognition (OCSR) engines to extract individual chemical components \cite{mcdaniel1992kekule,casey1993optical,ibison1993chemical,valko2009clide,filippov2009optical,frasconi2014markov,staker2019molecular,rajan2021decimer,yoo2022image,qian2023molscribe,chen2024molnextr}. Building on these efforts, work on reaction diagram parsing has emerged\cite{nguyen2020chemu,guo2021automated,wilary2021reactiondataextractor}.
However, these methods are often constrained by predefined rules or rigid heuristics, which struggle to generalize across unconventional layouts or noisy backgrounds.
The rapid proliferation of Vision Language Models (VLMs) \cite{qiu2025gated,zhao2023survey,chang2024survey,zhang2024vision,ghosh2024exploring} has recently shifted the paradigm toward learning-based methodologies \cite{qian2023rxnscribe,chen2025towards}. For instance, RxnIM \cite{chen2025towards} trains a multimodal VLM using a three-stage training strategy tailored to different chemical reaction image parsing tasks. 
Nevertheless, limited spatial layout understanding, confusion under intertwined visual elements, and insufficient integration of recognition and reasoning reveal the boundaries of VLMs in capturing the inherent complexity and heterogeneity of chemical graphics, making it difficult for them to handle the diverse illustrative styles and layout conventions in scientific literature. These limitations highlight three fundamental and yet unresolved challenges in reaction diagram parsing:

(1) \textbf{Visual and structural complexity:} Reaction diagrams are characterized by high-density interleaving of heterogeneous visual entities, including skeletal structures, varied arrow geometries, and textual descriptors. The primary difficulty lies in the intricate spatial geometries and non-linear topologies inherently present in scientific literature. Unlike simple linear transformations, complex diagrams often feature multi-step transformations and fragmented reaction sequences, where visual proximity does not consistently correlate with chemical connectivity. This makes reliable entity association and global relation modeling particularly challenging. (2) \textbf{Limited generalization:} Current learning-based methods often rely on dataset-specific visual or structural patterns. This results in a fragile generalisation capability when the model is confronted with the vast ``stylistic diversity''. Factors such as special artistic conventions (e.g., varying arrow head styles) and the emergence of novel, unconstrained diagrammatic formats create a significant distributional shift. Consequently, models trained on curated or synthetic benchmarks frequently suffer from catastrophic performance degradation when applied to previously unseen figures from diverse literature sources. (3) \textbf{Disjoint perception and reasoning:} Most existing reaction diagram parsing methods adopt a sequential pipeline that separates visual perception from chemical reasoning. This unidirectional design causes perception errors to propagate into later reasoning stages without correction, undermining global chemical consistency. Moreover, the lack of bidirectional interaction prevents the effective integration of spatial layout cues, semantic information, and chemistry-aware constraints, making it particularly difficult to recover implicit reaction semantics or handle highly ambiguous and complex diagrams.

\begin{figure*}[t]
\centering
\includegraphics[width=1\textwidth]{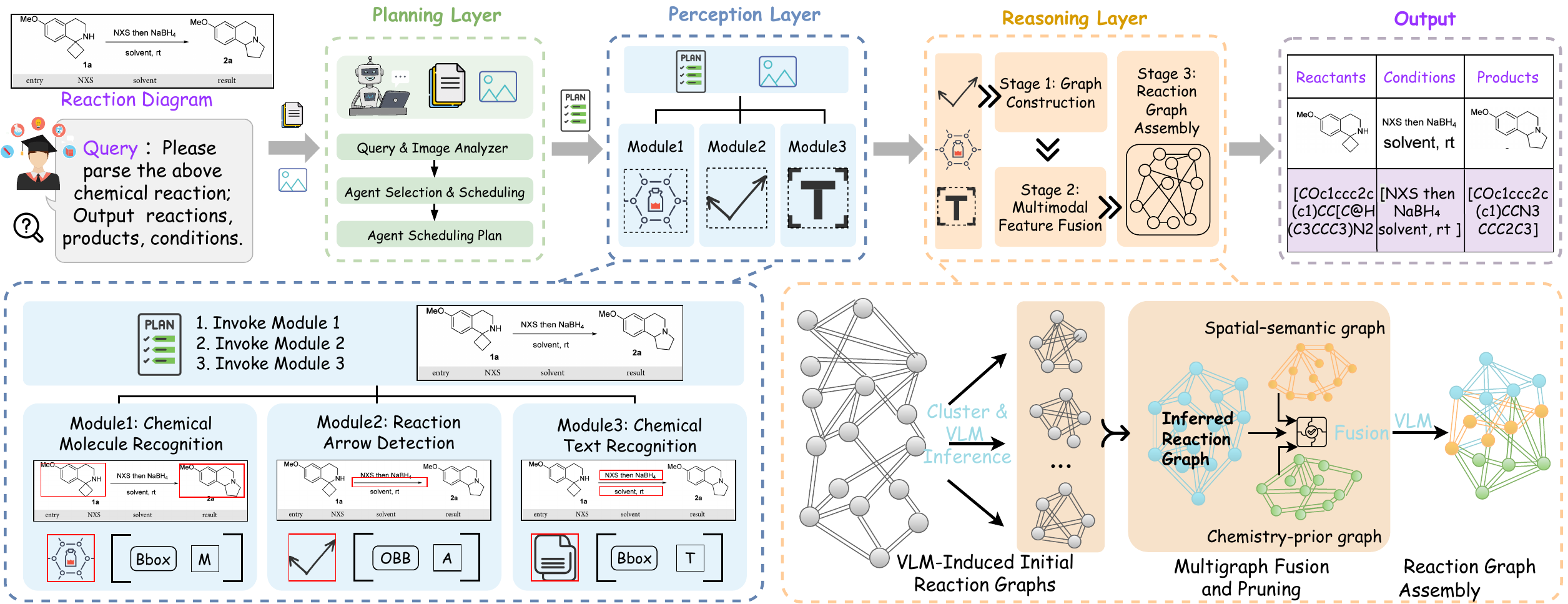}
\vspace{-0.2in}
\caption{Overview of \textbf{MACReD}, illustrating agent-level collaboration across planning, perception, and reasoning layers.}
\label{fig:framework}
\vspace{-0.1in}
\end{figure*}

To address these challenges, we introduce hierarchical multi-agent collaboration ~\cite{dorri2018multi,tran2025multi}, and propose \textbf{MACReD} (Multi-Agent Collaborative Reasoning for Reaction Diagram Parsing), a unified framework that tightly couples visual perception and chemical reasoning, as shown in Figure ~\ref{fig:framework}. Rather than treating reaction parsing as a monolithic end-to-end task, \textbf{MACReD} decomposes the process into coordinated agent-level subtasks across three hierarchical layers: \textbf{Planning, Perception, and Reasoning}.
Specifically, the Planning Layer dynamically orchestrates specialized agents according to the input query and diagram diagrammatic features, enabling flexible handling of diverse layouts. 
The Perception Layer employs dedicated agents for molecular localization, arrow understanding, and text extraction, producing structured and spatially grounded entities that reduce error propagation across heterogeneous visual elements. Building on these structured outputs, the Reasoning Layer performs global reaction reconstruction by explicitly modeling reactions as a shared reaction graph and integrating multiple sources of evidence. This graph serves as a unified intermediate representation that synthesizes heterogeneous cues, including spatial and structural relations, chemistry-aware semantic constraints, and VLM-induced reaction graphs. By fusing these multifaceted signals, \textbf{MACReD} ensures chemically valid inference while effectively resolving ambiguities in complex reaction topologies.

Our main contributions are summarized as follows:
\begin{itemize}[leftmargin=*]
  \item We propose \textbf{MACReD}, a multi-agent collaborative reasoning framework for reaction diagram parsing. By decomposing the complex parsing task into coordinated Planning, Perception, and Reasoning layers, MACReD coordinates specialized agents to jointly address visual complexity, limited generalization, and the disconnect between perception and chemical reasoning.
  \item We design a unified, graph-based fusion mechanism that synthesizes spatial structures, chemistry-aware constraints, and VLM-induced reaction entities. This integrated approach facilitates the robust and chemically consistent reconstruction of intricate, multi-step, and non-linear reaction pathways.
  \item Extensive experiments on benchmark datasets demonstrate that \textbf{MACReD} achieves state-of-the-art performance and strong generalization across diverse diagram layouts, including challenging multi-step and graph-structured reaction pathways.
\end{itemize}

\section{Related Work}
\label{sec:related}

\textbf{Vision Language Models and Multi-Agent Systems. }
Recent VLMs, including GPT~\cite{Hurst2024GPT4oSC}, Gemini~\cite{geminiteam2025geminifamilyhighlycapable}, and Qwen~\cite{Qwen3-VL}, have advanced cross-modal reasoning and scientific visual understanding. However, they remain limited in tasks requiring explicit structural modeling, fine-grained symbol grounding, and strict domain constraints, crucial for complex diagram interpretation. ChemVLM~\cite{li2025chemvlm} and Intern-S1-mini~\cite{bai2025interns1scientificmultimodalfoundation} improve molecular-level multimodal understanding via chemistry-aware pretraining, but as a single-model approach, it struggles with multi-step reaction diagrams. Intern-S1-mini~\cite{bai2025interns1scientificmultimodalfoundation} performs well on scientific benchmarks but lacks mechanisms for structured reaction reasoning. Meanwhile, multi-agent systems~\cite{talebirad2023multi,cheng2024exploring} offer an effective paradigm by decomposing tasks into agents. Recent studies in compositional image generation~\cite{li2025mccd}, workflow orchestration~\cite{shi2025flowxpert}, and zero-shot composed image retrieval~\cite{cheng2025generative} show the benefits of agent systems, motivating multi-agent architectures for reaction diagram parsing.

\textbf{Reaction Diagram Parsing. }
Early research focused on text-based extraction~\cite{nguyen2020chemu,guo2021automated}, using rule-based systems or pretrained language models to identify reaction components from natural language, which overlooked the rich knowledge encoded in graphical diagrams. Diagram-based systems, such as ReactionDataExtractor(RDE)~\cite{wilary2021reactiondataextractor,wilary2023reactiondataextractor}, initially relied on handcrafted geometric heuristics, but these proved notoriously brittle under complex layouts. Modern learning-based approaches, notably RxnScribe~\cite{qian2023rxnscribe}, adopted end-to-end sequence generation for greater robustness. However, such methods often entangle perception and reasoning, making it difficult to enforce explicit structural constraints or recover from local errors. Recent VLM-based methods~\cite{chen2025towards} leverage multimodal pretraining for reaction diagram understanding, yet they still lack explicit spatial modeling and chemistry-aware consistency enforcement. In contrast, our approach explicitly separates perception and reasoning into coordinated stages within a hierarchical framework, employing a unified graph representation to ensure both structural and chemical consistency.

\begin{figure}[!ht]
    \centering
    \includegraphics[width=0.75\linewidth]{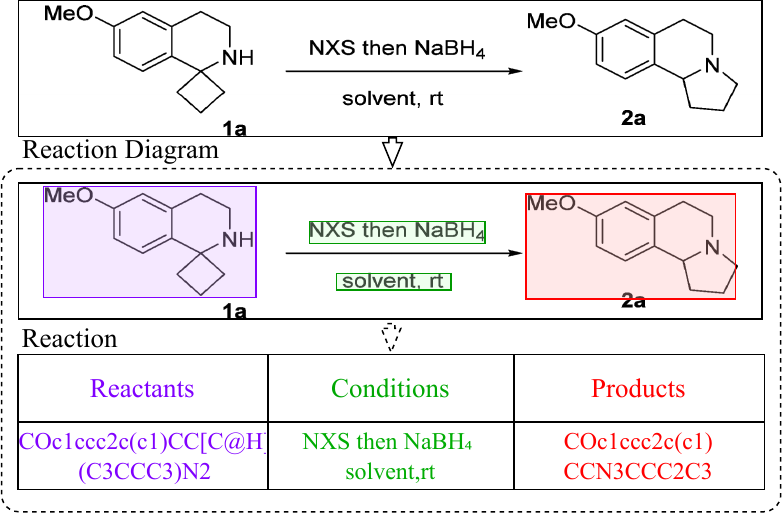}
    \vspace{-0.05in}
    \caption{Example of the reaction diagram parsing task.}

    \label{fig:problem}
    \vspace{-0.05in}
\end{figure}

\section{Method}
\label{sec:method}

\subsection{Problem Definition}
Reaction diagram parsing is a fundamental task in scientific document understanding that aims to automatically recover structured chemical reaction information from diagrammatic depictions. As illustrated in Figure ~\ref{fig:problem}, given a reaction diagram image \(D\), which may depict isolated elementary transformations or multi-step synthetic routes with complex and heterogeneous spatial layouts, the objective is to extract all reactions together with their corresponding reactants, products, and reaction conditions. Formally, reaction diagram parsing is formulated as learning a mapping function:
{\setlength{\abovedisplayskip}{3pt}
	\setlength{\belowdisplayskip}{3pt}
\begin{equation}
\mathcal{F}: D \mapsto R = \{R_1, R_2, \dots\},
\label{eq:task_mapping}
\end{equation}}\noindent

\noindent where the cardinality of \(R\) and the structure of each reaction are unknown and must be jointly inferred from the diagram. Each reaction \(R_i\) is represented as $R_i = (S_i, C_i, T_i)$, where \(S_i\), \(C_i\), and \(T_i\) denote the sets of reactants, conditions, and products, respectively.

Each reactant or product may be manifested through heterogeneous visual or symbolic entities, including molecular structure drawings, textual labels, symbolic identifiers, or their combinations, complicating reliable entity grounding and association. The set \(C_i\) may be empty when reaction conditions are implicit or omitted in the diagram, reflecting common practices in literature. Although practical reaction diagrams often contain drawing inconsistencies, OCSR noise, or incomplete annotations, chemically valid reactions approximately satisfy fundamental conservation constraints:

{\setlength{\abovedisplayskip}{-5pt}
	\setlength{\belowdisplayskip}{3pt}
\begin{equation}
\sum\nolimits_{r \in S_i} \mathbf{a}(r) = \sum\nolimits_{p \in T_i} \mathbf{a}(p), \quad
\sum\nolimits_{r \in S_i} \mathbf{q}(r) = \sum\nolimits_{p \in T_i} \mathbf{q}(p),
\label{eq:conservation}
\end{equation}}\noindent

\noindent where \(\mathbf{a}(\cdot)\) and \(\mathbf{q}(\cdot)\) denote atom-count vectors and formal charges, respectively. These constraints are not assumed to strictly hold in the raw input diagrams, but instead serve as weak global regularities that guide downstream reasoning and help resolve ambiguities in structured reaction inference.

To address the challenges of accurately parsing complex chemical reaction diagrams, we propose \textbf{MACReD}, a hierarchical framework driven by VLM for collaborative chemical diagram understanding. As illustrated in Figure ~\ref{fig:framework}, \textbf{MACReD} is organized into three hierarchical layers, namely Planning, Perception, and Reasoning.
Each agent is guided by task-specific modules and carefully customized system prompts, enabling adaptive and coordinated collaboration across visual perception and symbolic reasoning stages.
Through this layered and agent-centric design, \textbf{MACReD} achieves robust and highly flexible interpretation of complex, multi-entity reaction diagrams.
All prompt templates can be found in Appendix ~\ref{sec:promptdesign}.

\subsection{Planning Layer}

The Planning Layer functions as the top-level control module of \textbf{MACReD}, responsible for global task decomposition and coordination among specialized agents. It resolves issues stemming from high task heterogeneity, uncertain reasoning paths, and the rigidity of fixed pipelines. The Planning Agent operates as a multimodal controller that jointly analyzes the user query and the visual content of the input reaction diagram, and dynamically determines an adaptive sequence of downstream agent invocations.

Rather than relying on explicit search-based planning or hand-crafted utility maximization, the Planning Agent formulates decision making as a context-conditioned agent routing problem. Its role is to dynamically select and order specialized agents according to task differences and diagram complexity, spatial organization, and semantic uncertainty, enabling adaptive and scalable reasoning over heterogeneous reaction diagrams. Let
\(
\mathcal{A} = \{A_1, A_2, \dots, A_m\}
\)
denote the set of available agents. 
The Planning Agent predicts an ordered execution plan: 
\(
\pi = \{\alpha_{1}, \alpha_{2}, \dots\}, \alpha_{k} \in \mathcal{A}
\), where each element corresponds to invoking a specific agent at step \(k\). Unlike action-level planning, the plan operates at the level of agent selection, defining a high-level control flow for the system.
Agent selection is performed autoregressively via a multimodal routing function:
\(
\alpha_k = f_{\mathrm{plan}}({F}_D, Q, \mathcal{C}_k),
\)
where \({F}_D\), \(Q\), and \(\mathcal{C}_k\) denote diagram-level visual features, task query, and an evolving context that aggregates intermediate evidence and prior agent decisions. In practice, this routing is realized through lightweight prompt-based multimodal inference, enabling flexible and efficient agent coordination.

\begin{figure}[t]
\centering
\includegraphics[width=0.4\textwidth]{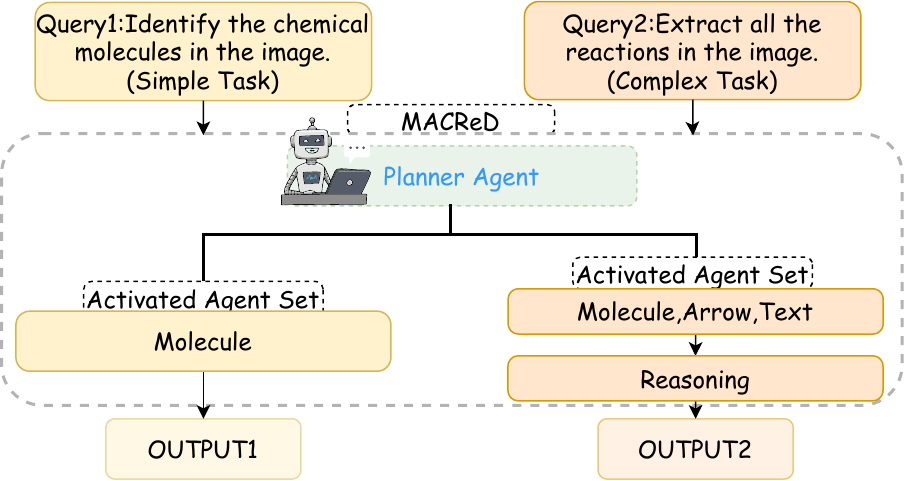}
\vspace{ -0.1in}
\caption{Example of routing mechanism.}
\label{fig:planner}
\vspace{-0.25in}
\end{figure}

As illustrated in Figure ~\ref{fig:planner}, this adaptive routing mechanism enables \textbf{MACReD} to flexibly orchestrate different subsets of agents according to task requirements. 
For multi-step reaction pathways, the Planning Agent activates all relevant agents in a coordinated sequence, while simpler tasks invoke only a minimal subset of agents, improving computational efficiency.

This hierarchical control strategy enables the Planning Layer to scale across task granularities, from molecule-level perception to multi-step synthesis reconstruction, while maintaining robust multimodal integration.

\subsection{Perception Layer}

To address the prominent visual and structural complexity and limited generalization in reaction diagram parsing, the Perception Layer assumes the core of comprehensively understanding both the visual and textual content of reaction diagrams,

serving as the bridge between raw image pixels and symbolic chemical representations. This layer decomposes the diagram perception task into a set of specialized agents, each focusing on a specific visual modality and collaboratively transforming heterogeneous inputs into structured entities suitable for downstream reasoning.
Specifically, the Molecule Agent detects molecular structure depictions, while the Molecule-Recognition Agent converts cropped molecular regions into SMILES; the Arrow Agent identifies and classifies reaction arrows; the Text Agent together with the Text-Recognition Agent extracts and normalizes textual annotations. These modular agents work collaboratively to produce a unified set of structured entities, providing a robust foundation for subsequent reasoning.

Given an input image \(D\), the Perception Layer outputs a unified set of structured entities: 
\(
V = \{v_i\}_{i=1}^{N}
\), 
\(
v_i = (b_i, \tau_i, \rho_i)
\),

where the entity representation consists of three components: 
the spatial region \(b_i\), the semantic type \(\tau_i \in \{\text{molecule}, \text{arrow}, \text{text}\}\), 
and the associated semantic information \(\rho_i\). 
Together, these components form a spatially grounded and semantically normalized representation that provides the basis for subsequent reaction graph construction.

\textbf{Molecule Agents and Molecule Recognition. }
Understanding molecular structures in reaction diagrams exceeds the capabilities of conventional object detection and OCSR, due to diverse drawing styles, overlapping annotations, low scan resolution, and frequent deviations from modern cheminformatics conventions. Challenges such as blurred bonds, inconsistent atom labels, and occlusion by arrows or text often lead to errors in end-to-end recognition.

To address these challenges, we adopt a decoupled two-stage perception pipeline, explicitly separating spatial localization from structural interpretation. First, a CNN-based detection module performs object-level molecular localization to generate preliminary inputs for the agents. The Molecule Agent refines these results, isolating molecular regions from complex backgrounds.  
Next, the Molecule-Recognition Agent leverages MolScribe and RDKit based image2graph and graph2smiles modules to convert cropped molecular regions into SMILES or molecular graphs. VLM
are subsequently applied to perform chemical semantic and commonsense reasoning, such as atom types and stereochemistry, to correct potential recognition errors.
This factorization avoids entangling geometric localization with graph reconstruction errors, enabling more robust recovery of molecular topology under noisy or incomplete visual inputs. Empirically, this modular design improves generalization across heterogeneous diagram styles and facilitates downstream error correction when chemical constraints are violated.

\textbf{Arrow Agent. }
Reaction arrows serve as the primary carriers of essential causal and directional semantics in reaction diagrams, yet their visual realization is highly unconstrained in practice. Arrows may be curved, tilted, multi-headed, partially occluded, or arranged in intricate complex multi-step layouts, rendering heuristic-based or axis-aligned detection strategies unreliable.

To address this, the Arrow Agent performs geometric localization and semantic classification of arrows. An Oriented Bounding Box (OBB) detection module, trained on fine-grained geometric annotations from the RxnScribe dataset, captures arrows of diverse shapes and orientations. Beyond detection, the module classifies arrows into functional categories (e.g., forward, reversible, resonance), providing explicit cues for reaction directionality and transformation semantics. To enhance robustness, the VLM combines the outputs of the module with visual semantic information for reasoning, to cross-check and correct arrow classifications.

\textbf{Text Agents and Text Recognition. }

Textual annotations in reaction diagrams encode critical contextual information such as reagents, catalysts, solvents, temperatures, and yields. However, chemical text extraction is substantially more challenging than generic OCR due to domain-specific symbols, subscript notation, abbreviations, and frequent OCR confusion between visually similar tokens (e.g., ``Cl'' vs. ``CI'').

Text detection module first identifies all text regions, which are carefully refined by the Text Agent. A PaddleOCR~\cite{cui2025paddleocr30technicalreport} based recognition module performs initial transcription, followed by normalization by the Text-Recognition Agent. Synonymous expressions (e.g., ``$FeCl_3$'' vs. ``ferric chloride'') are unified and mapped to structured chemical entities. By explicitly modeling text recognition as a normalization task rather than pure transcription, the system effectively achieves higher robustness under noisy scans and heterogeneous annotation styles, while aligning textual information with molecular structures to support accurate global reaction diagram construction through multi-agent collaboration.

The introduction of specialized perception modules provides precise and structured evidence to support agent level reasoning, while the high level reasoning capability of VLM further compensates for the limitations of lightweight models under complex, diverse, and non-canonical real-world scenarios. Together, this collaborative design substantially enhances the system's robustness and generalization on challenging reaction diagrams, establishing a solid foundation for subsequent Global Reaction Inference.

\subsection{Reasoning Layer}

To integrate perception and reasoning, three intermediate graphs, the Spatial--Semantic, Chemistry-Aware, and VLM-Induced Reaction Entity graphs, propagate local spatial and semantic cues for context-aware relation scoring beyond simple pairwise heuristics.
These graphs are then integrated into a unified representation through a Multigraph Fusion mechanism.
Built upon this fused graph, the Reaction-Combiner Agent leverages symbolic chemical rules together with learned reaction patterns to iteratively refine, validate, and consolidate candidate reactions, assembling complete and coherent reaction equations. By jointly integrating multimodal graph evidence, chemistry-aware constraints, and preliminary reaction selections from a VLM, the agent performs global reasoning to produce consistent and chemically valid reaction equations.

\textbf{Spatial-Semantic Graph Construction.}
We adopt a lightweight GCN~\cite{wu2020comprehensive, zhang2019graph} not for deep relational reasoning, but to propagate local spatial cues across neighboring entities, enabling context-aware relation scoring beyond pairwise heuristics: \(G_{\text{space}} = (V, E_{\text{space}})\),

where edges are inferred jointly from spatial proximity and visual-semantic coherence. Message passing is performed as
{\setlength{\abovedisplayskip}{0pt}
	\setlength{\belowdisplayskip}{0pt}
\begin{equation}
\mathbf{h}_i^{(\ell+1)} =
\sigma\!\left(
\sum_{j \in \mathcal{N}(i)}
\mathbf{W}_1 \mathbf{h}_j^{(\ell)}
+
\mathbf{W}_2 \mathbf{e}_{ij}
\right),
\quad
e_{ij} \in E_{\text{space}},
\label{eq:gcn_update}
\end{equation}}\noindent

\noindent where \(\mathbf{h}_i^{(\ell)}\), \(\mathbf{e}_{ij}\), and \(\sigma(\cdot)\) respectively correspond to the layer-wise node representations, the relative spatial and visual-semantic features entity pairs, and the nonlinear activation function. 
This stage encodes layout-aware and visually grounded interactions among entities, yielding an initial spatial dependency graph.

\textbf{Chemistry-Aware Graph Construction.}
To incorporate domain level chemical constraints and suppress physically implausible relations, we construct a chemistry-aware graph: \(G_{\text{chem}} = (V, E_{\text{chem}})\), where \(V\) and \(E_{\text{chem}}\) denote entities and chemically plausible edges

Chemical affinity between entity pairs is quantified using molecular fingerprint similarity and charge consistency: \(s_{ij}^{\text{fp}} = \mathrm{Tanimoto}(\mathrm{FP}_i, \mathrm{FP}_j)\), \(\Delta q_{ij} = \lvert q_i - q_j \rvert\).

Here, \(s_{ij}^{\text{fp}}\) and \(\Delta q_{ij}\) respectively correspond to the fingerprint-based similarity between entities \(i\) and \(j\) and the absolute difference in their formal charges.
These two factors are integrated via a lightweight chemistry-aware scoring function: \(e_{ij}^{\text{chem}} =
f_{\text{chem}}\!\left(
s_{ij}^{\text{fp}},
\Delta q_{ij}
\right)\), where \(f_{\text{chem}}(\cdot)\) is the aggregation function and \(e_{ij}^{\text{chem}}\) the resulting plausibility score.

where higher scores indicate stronger chemical plausibility.
Edges exhibiting sufficient chemical consistency are retained through thresholding: \(E_{\text{chem}}
=
\left\{
(i,j)\ \middle|\ e_{ij}^{\text{chem}} > \tau_{\text{chem}}
\right\}\),

where \(\tau_{\text{chem}}\) specifies a pruning threshold that controls the inclusion of chemically plausible relations.
These chemistry-aware scores act as soft inductive biases rather than hard constraints, allowing uncommon transformations to be preserved during subsequent graph fusion and reasoning.

\textbf{VLM-Induced Reaction Entities Graph.}
To incorporate high-level semantics that are difficult to capture via local spatial or chemical cues, we first generate a preliminary reaction entities graph induced by a VLM, which serves as an intermediate graph rather than a final prediction.
Specifically, the full vertex set is partitioned into localized subgraphs: \(\{V_1, V_2, \dots\} =
\mathrm{Cluster}(V)\),

thereby constraining the contextual scope and reducing combinatorial complexity. Without such localization, VLM predictions tend to conflate unrelated reaction steps in densely populated diagrams, leading to spurious global associations.
Each subgraph is then processed independently by the VLM to infer potential candidate reaction relations: \(G_{\text{VLM}} = \{G_1, G_2, \dots\}\), \(G_k = (V_k, E_k^{\text{VLM}})\),

where \(G_{\text{VLM}}\), \(G_k\), and \(E_k^{\text{VLM}}\) respectively represent the collection of VLM-induced subgraphs, an individual subgraph associated with entity subset \(V_k\), and the set of reaction relations inferred within that subset.
Edges correspond to predicted directional reaction relations: \(E_k^{\text{VLM}} =
\left\{
(v_i, v_j, r)\ \middle|\
v_i \xrightarrow{r} v_j
\right\}\),

where \((v_i, v_j, r)\) respectively indicate the source, target entity, and the inferred reaction type associated with a directed relation.

Unlike the spatial-semantic and chemistry graphs, which primarily emphasize local consistency cues, the VLM-induced graphs provide coarse, semantically informed reaction hypotheses. These hypotheses are subsequently fused with complementary evidence streams and refined by the downstream reasoning module to yield the final reaction interpretation.

\textbf{Multigraph Fusion and Pruning.}
The spatial-semantic graph, chemistry-prior graph, and VLM-induced hypotheses are consolidated into a unified reasoning graph: \(G_{\mathrm{fuse}} = (V, E_{\mathrm{fuse}})\),

where \(V\) and \(E_{\mathrm{fuse}}\) respectively correspond to the shared set of entities and the set of candidate relations obtained after evidence fusion.
Each candidate edge is assigned a fused confidence score via weighted aggregation of multiple evidence sources:
{\setlength{\abovedisplayskip}{3pt}
	\setlength{\belowdisplayskip}{3pt}
\begin{equation}
s_{ij}^{\mathrm{fuse}}
=\alpha_{\mathrm{space}}\, s_{ij}^{\mathrm{space}}+\alpha_{\mathrm{chem}}\, s_{ij}^{\mathrm{chem}}+\alpha_{\mathrm{init}}\, s_{ij}^{\mathrm{init}},
\label{eq:fuse_score}
\end{equation}}\noindent

\noindent where \(s_{ij}^{\mathrm{space}}\), \(s_{ij}^{\mathrm{chem}}\), and \(s_{ij}^{\mathrm{init}}\) respectively reflect spatial-semantic evidence, chemistry-prior consistency, and VLM-induced reaction hypotheses. Here, \(\alpha_{\ell}\) controls their relative contributions under a normalized weighting scheme, and there is $\sum\nolimits_{\ell} \alpha_{\ell} = 1$.
Edges with low aggregated confidence are pruned: \(E_{\mathrm{fuse}}
=
\left\{
(i,j)\ \middle|\ s_{ij}^{\mathrm{fuse}} > \tau_{\mathrm{fuse}}
\right\}\),

where \(\tau_{\mathrm{fuse}}\) specifies a pruning threshold that filters out weakly supported relations.
This fusion-and-pruning process yields a sparse and noise-suppressed reasoning graph that retains only relations jointly supported by multiple complementary evidence sources.

\textbf{Global Reaction Inference.}
Candidate configurations are enumerated only over connected components of the pruned fused graph, whose sparsity significantly limits the combinatorial space in practice. The optimal reaction structure is selected by maximizing the accumulated structural support across all constituent edges: \(R^{*}
=
\arg\max_{R_i}
\sum_{(v_j, v_k) \in E(R_i)}
s_{jk}^{\mathrm{fuse}}\),

where \(R_i\), \(E(R_i)\), and \(s_{jk}^{\mathrm{fuse}}\) respectively correspond to a candidate reaction entity, its associated set of reaction edges, and the fused confidence scores.

Throughout the reasoning process, a shared multimodal context is continuously maintained and updated as a global state, aggregating visual evidence, linguistic cues, and chemistry-aware semantic information. This hierarchical reasoning pipeline, comprising spatial, semantic propagation, chemistry-aware pruning, VLM-guided hypothesis generation, and multimodal evidence fusion, produces a sparse and interpretable reaction graph that preserves chemically meaningful relations while effectively avoiding the combinatorial explosion associated with fully connected formulations.

\section{Experiments}
\label{sec:experiments}

\begin{figure}[t]
\centering
\includegraphics[width=0.47\textwidth]{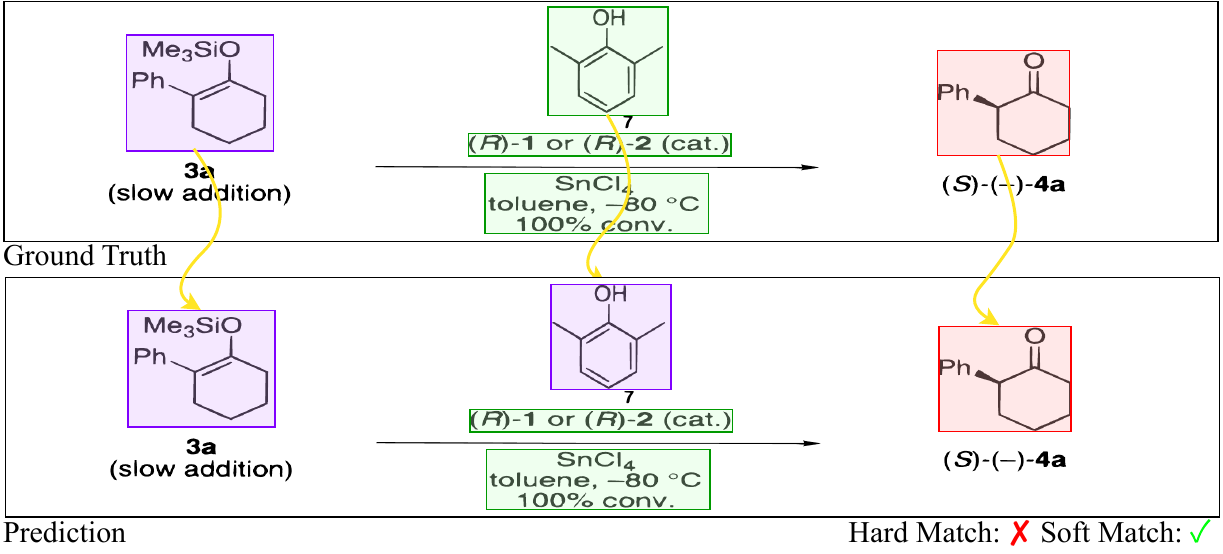}
\vspace{-0.1in}
\caption{\textbf{Hard Match} and \textbf{Soft Match} evaluation protocols. }

\label{fig:hard_soft_match}
\vspace{-1.2em}
\end{figure}

\subsection{Datasets and Evaluation Metrics}

\textbf{Datasets and Preprocessing.}
We evaluate the proposed \textbf{MACReD} framework on the RxnScribe benchmark dataset, which is currently one of the most comprehensive publicly available resources for reaction diagram understanding. 
To facilitate more accurate structural parsing, we further re-annotated the dataset with fine-grained arrow annotations using OBB, which enables robust detection of complex arrow geometries.

\textbf{Baselines.}
We compare \textbf{MACReD} against two categories of baseline methods.
General-purpose VLM: Closed-source: GPT-4o~\cite{Hurst2024GPT4oSC}, Gemini-2.5-Pro~\cite{geminiteam2025geminifamilyhighlycapable}, QwenVL-Max~\cite{Qwen3-VL}; Open-source: Qwen2.5-VL-7B~\cite{bai2025qwen25vltechnicalreport}, InternVL3-8B~\cite{zhu2025internvl3exploringadvancedtraining}, MiMo-VL-7B~\cite{xiaomi2025mimo}; Chemistry: ChemVLM-8B~\cite{li2025chemvlm}, Intern-S1-mini~\cite{bai2025interns1scientificmultimodalfoundation}.  
Task-specific systems: Rule-based: RDE~\cite{wilary2021reactiondataextractor}, RDE2~\cite{wilary2023reactiondataextractor}; Learning-based pipelines: OChemR~\cite{M.Martori2022}; End-to-end approaches: RxnScribe~\cite{qian2023rxnscribe}, RXNIM~\cite{chen2025towards}.

See Appendix B for more details.

\textbf{Evaluation Metrics.}
Evaluating presents unique challenges, ground-truth annotations and predictions are represented as sets of structured reactions. Exact one-to-one correspondence is often infeasible due to small bounding-box variations, entity segmentation inconsistencies, or differing reaction orderings. To address these ambiguities, we use Hard Match and Soft Match, shown in Figure ~\ref{fig:hard_soft_match}.

For a given diagram, the ground-truth set is denoted as \(G = \{ R_1, R_2, \dots, R_n \}\), and the predicted set as \(P = \{ \hat{R}_1, \hat{R}_2, \dots, \hat{R}_m \}\). 
Each reaction \(R_i\)(or \(\hat{R}_i\)) is represented as a triplet \((S_i, C_i, T_i)\). For each entity in \(R_i\), we identify the entity in \(\hat{R}_i\) with the Intersection over Union (IoU) of bounding boxes. If the IoU exceeds a threshold of 0.5, the two entities are considered successfully matched.

Hard Match requires all reactants, conditions, and products to match, while Soft Match considers only molecular entities, ignoring text. This relaxation mitigates annotation ambiguity and focuses on the molecular relationship extraction. As shown in Figure \ref{fig:hard_soft_match}, Hard Match evaluates full structural fidelity; Soft Match captures molecular correctness under annotation variability.
For both Hard Match and Soft Match, we compute Precision, Recall, and F1-score.

\vskip -2em
\subsection{Main Results}
\textbf{Reaction Extraction Performance.}
Table~\ref{tab:reaction_results} summarizes the comparative performance of different methods on reaction extraction under both Hard Match and Soft Match evaluation protocols. Reaction diagram parsing remains highly challenging, with rule-based systems and single-model approaches struggling to achieve globally consistent and accurate extraction. In contrast, \textbf{MACReD}'s hierarchical multi-agent design delivers significant improvements in both structural and semantic accuracy, consistently outperforming all baselines across Precision, Recall, and F1 metrics, and demonstrating superior structural fidelity and robust semantic understanding.

Classical rule-based systems, such as RDE and OChemR, exhibit very limited performance, with Hard Match F1 scores below 10\%.
This limitation primarily stems from their reliance on deterministic heuristics and handcrafted spatial rules, which struggle to handle complex diagrams in real-world chemical literature.
The enhanced RDE 2.0 achieves noticeably higher performance by incorporating a deep learning component detection.
However, its continued dependence on rigid entity association rules restricts its ability to robustly reconstruct complete and globally consistent reaction structures.

Learning-based approaches, including RxnScribe and RxnIM, demonstrate substantial performance gains, highlighting the effectiveness of data-driven models in capturing complex visual and semantic patterns.
Nevertheless, their performance remains constrained by limited global reasoning capabilities and insufficient integration of heterogeneous evidence.
In complex or highly variable reaction diagrams, such as those involving multiple reactants or products, branched reaction arrows, or reactants positioned above arrows, these methods often exhibit partial correctness, successfully identifying some components while failing on others.
This behavior reflects the intrinsic diversity and structural complexity of reaction diagrams, combined with the limited scale and coverage of available annotated training data.
When confronted with diagram structures that deviate from the training distribution, these models reveal clear deficiencies in generalization.

Open-source VLMs fail to produce valid reaction extraction results under both evaluation protocols.
All evaluated models score zero under both Hard and Soft Match criteria.
Although these models perform well on general multimodal understanding tasks, they are unable to reconstruct structured reaction graphs, even specialized chemical models such as ChemVLM fail completely.
This indicates that reaction diagram understanding is a complex structured knowledge extraction problem, rather than a mere multimodal generation task.
The difficulty of this task exceeds the capability limits of small-scale VLMs, and even domain-specific fine-tuning is insufficient to provide reliable structural reconstruction ability.

\begin{table}[t]

  \caption{Evaluation of model performance in reaction diagram parsing.}
  \vspace{-0.8em}
  \label{tab:reaction_results}
  \centering
  \begin{tabular}{@{}lcccccc@{}}
    \toprule
     & \multicolumn{3}{c}{Hard Match} & \multicolumn{3}{c}{Soft Match} \\ 
    \cmidrule(lr){2-4} \cmidrule(lr){5-7}
    Model & Prec. & Recall & F1 & Prec. & Recall & F1 \\
    \midrule
    \multicolumn{7}{c}{\textbf{Open-source Models}} \\
    \midrule
    Qwen2.5-VL-7B  & 0.0 & 0.0 & 0.0  & 0.0 & 0.0 & 0.0 \\
    InternVL3-8B & 0.0 & 0.0 & 0.0  & 0.0 & 0.0 & 0.0 \\
    MiMo-VL-7B & 0.0 & 0.0 & 0.0  & 0.0 & 0.0 & 0.0 \\
    ChemVLM-8B & 0.0 & 0.0 & 0.0  & 0.0 & 0.0 & 0.0 \\
    Intern-S1-mini & 0.0 & 0.0 & 0.0  & 0.0 & 0.0 & 0.0 \\
    
    \midrule
    \multicolumn{7}{c}{\textbf{Close-source Models}} \\
    \midrule
    Gemini-2.5-Pro & 0.0 & 0.0 & 0.0 & 23.3 & 24.5 & 23.9 \\
    GPT4o & 0.5 & 0.1 & 0.2 & 2.3 & 1.9 & 2.1 \\
    QwenVL-Max & 1.0 & 0.5 & 0.7 & 6.4 & 6.1 & 6.2 \\
    
    \midrule
    \multicolumn{7}{c}{\textbf{Trained Methods}} \\
    \midrule    RDE & 4.1 & 1.3 & 1.9 & 19.4 & 5.9 & 9.0 \\
    RDE2.0 & 42.1 & 41.8 & 42.0 & 49.5 & 49.1 & 49.4 \\
    OChemR & 4.4 & 2.8 & 3.4 & 12.4 & 7.9 & 9.6 \\
    RxnIM & 74.9 & 70.1 & 72.4 & 81.9 & 77.3 & 79.5 \\
    RxnScribe & 72.3 & 66.2 & 69.1 & 83.8 & 76.5 & 80.0 \\
    
    \midrule
    \multicolumn{7}{c}{\textbf{Ours}} \\
    \midrule
    MACReD(GPT) & 75.7 & 73.1 & 74.4 & 86.8 & \textbf{82.3} & 84.5 \\
    \textbf{MACReD(Gemini)} & \textbf{76.9} & \textbf{73.6} & \textbf{75.2} & \textbf{87.2} & 82.1 & \textbf{84.6} \\
    \bottomrule
  \end{tabular}
  \vspace{-0.2in}
\end{table}

In contrast, closed-source VLMs exhibit slightly better but still limited performance on this specialized task.
Their F1 remains near zero under the Hard Match criterion and does not exceed 7\% under Soft Match.
These results further suggest that, even at larger-scales model, generic multimodal reasoning alone is insufficient for accurate chemical reaction diagram parsing.
This indicates that the challenge is not merely a matter of model size, but instead exceeds the capability boundary of a single monolithic model, necessitating the incorporation of auxiliary mechanisms or system-level solutions, such as specialized recognition modules, multi-agent collaboration.

Our framework achieves the best performance among all methods, reaching 75.2\% F1 under Hard Match and 84.6\% under Soft Match. 
Compared with the SOTA method, it improves F1 by +6.1\% under Hard Match and +4.6\% under Soft Match. 
These improvements can be attributed to the proposed hierarchical multi-agent architecture, which explicitly decomposes the task into specialized perception and reasoning stages.
By coordinating molecule, arrow, and text agents with the Reaction-Combiner Agent through a context-aware reasoning pipeline, \textbf{MACReD} effectively aligns molecular entities, reaction directionality, and textual semantics.

When confronted with complex tasks characterized by diverse layouts, variable drawing conventions, and densely overlapping textual annotations, a single VLM exhibits significant limitations in capturing domain-specific chemical structures and constraints. This exposes a fundamental gap between its capabilities and the requirements for extracting scientific knowledge from complex, structure-rich diagrams. Therefore, integrating multi-agent perception with structured, graph-based reasoning is essential for achieving accurate and generalizable reaction extraction.

\begin{figure}[t]
    \centering
    \includegraphics[width=1\linewidth]{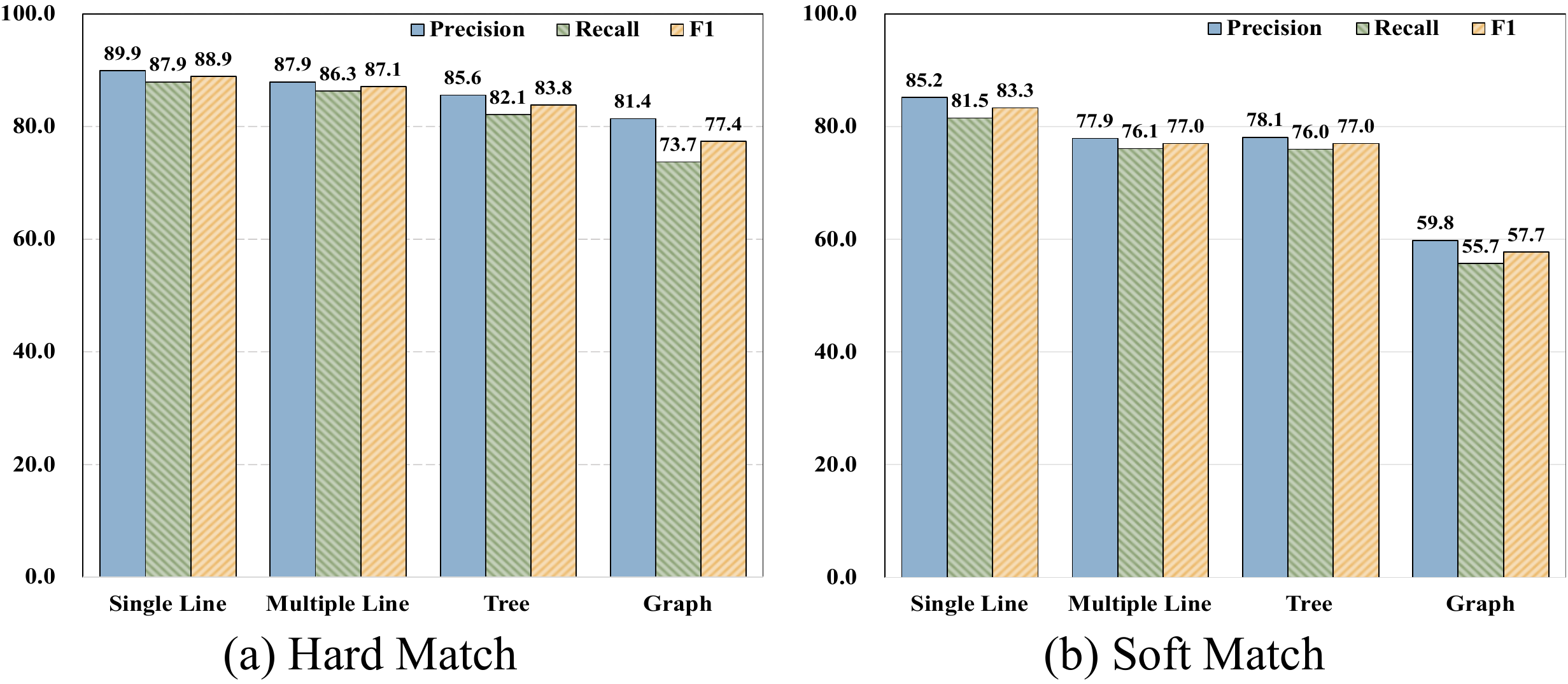}
    \vspace{-0.2in}

    \caption{\textbf{MACReD}'s reaction diagram parsing performance on four layouts.}
    \label{fig:type_results}
    \vspace{-0.6cm}
\end{figure}

\textbf{Structural Representation Comparison.}
MACReD demonstrates robust and consistent reaction extraction across diverse diagram layouts, as shown in Figure~\ref{fig:type_results}, which presents a detailed breakdown of its performance across four representative diagram types commonly found in chemical literature. These layouts correspond to increasing levels of structural complexity and relational ambiguity, providing a systematic evaluation of the model's robustness to diverse diagrammatic organizations.

\textbf{MACReD} achieves its strongest performance on Single Line diagrams, where reactions follow a predominantly linear topology with minimal branching and unambiguous arrow-molecule associations. This setting aligns well with the model's perception and reasoning pipeline, allowing accurate entity grounding and reaction graph construction.
Performance degrades moderately for Multiple Line and Tree layouts, which introduce vertically stacked reactions or branching intermediates that increase spatial overlap and reaction-level ambiguity. In these cases, \textbf{MACReD} maintains competitive accuracy, indicating its ability to resolve non-linear dependencies and shared intermediates through graph-based evidence fusion.
Graph-structured diagrams pose the greatest challenge, as they exhibit dense connectivity, non-linear reaction flow, and multiple plausible reaction groupings. These characteristics substantially increase the combinatorial complexity of reaction assignment, leading to lower performance. Notably, the gap between Hard and Soft Match remains controlled, suggesting that most errors arise from structural association rather than molecular misrecognition.

Despite the increasing difficulty across layouts, \textbf{MACReD} demonstrates relatively stable degradation trends, highlighting the effectiveness of its unified reaction graph representation and multi-agent reasoning mechanism. By jointly integrating spatial cues, chemical constraints, and high-level semantic, the framework remains capable of producing chemically coherent reaction reconstructions even under highly complex and densely connected diagram structures.

\textbf{Summary of Findings.} 
In summary, \textbf{MACReD} achieves SOTA performance on reaction extraction under both Hard and Soft Match criteria, consistently surpassing all baseline models. Notably, the framework demonstrates robust performance across diverse structural layouts, achieving the best score on Single Line diagrams, while maintaining competitive results even on available Graph layouts.
Preserving structural fidelity and chemical consistency across heterogeneous reaction types. The hierarchical, multi-agent design allows cooperative integration of molecular, textual, and graphical information, providing strong adaptability to various diagrammatic configurations ranging from linear sequences to complex multi-branch graphs.
Overall, these results confirm the effectiveness, robustness, and versatility of \textbf{MACReD} in unified visual and chemical reasoning, highlighting the critical role of multi-agent cooperation and context-aware inference in achieving accurate and chemically coherent reaction reconstruction.

\subsection{Ablation Studies}

To quantify the contribution of individual components in \textbf{MACReD}, we conducted ablation studies under both Hard and Soft Match (Table~\ref{tab:ablation}), which reveal that its performance gains arise from the strongly synergistic interaction between perception accuracy, semantic grounding, and graph-based reasoning. Components responsible for structural constraints (arrow understanding and multigraph fusion) have the greatest impact on Hard Match performance, while semantic modules (molecule and text recognition) are essential for maintaining chemical validity under both evaluation criteria. These results empirically support the design of \textbf{MACReD} as a tightly coupled multi-agent system rather than a loosely connected perception pipeline, demonstrating that accurate and chemically coherent reaction graph reconstruction depends on complementary contributions from all modules.

\begin{table}[!ht]
  \caption{Ablation analysis of key system components.}
  \label{tab:ablation}

  \centering
  \small
  \setlength{\tabcolsep}{3pt}
  \begin{tabular}{@{}lcccccc@{}}
    \toprule
     & \multicolumn{3}{c}{Hard Match} & \multicolumn{3}{c}{Soft Match} \\ 
    \cmidrule(lr){2-4} \cmidrule(lr){5-7}
      & Prec. & Rec. & F1 & Prec. & Rec. & F1 \\
    \midrule
    \textbf{MACReD} & \textbf{76.9} & \textbf{73.6} & \textbf{75.2} & \textbf{87.2} & \textbf{82.1} & \textbf{84.5} \\
    w/o Arrow & 38.6 & 32.7 & 35.4 & 65.0 & 55.2 & 59.7 \\ 
    w/o Molecule-Recognition & 34.7 & 32.2 & 33.4 & 62.1 & 57.7 & 59.8 \\
    w/o Text-Recognition & 51.9 & 48.6 & 50.2 & 73.7 & 69.0 & 71.3 \\
    w/o Multigraph-Fusion & 52.3 & 48.7 & 50.4 & 64.4 & 56.8 & 60.4 \\
    w/o Post-Processing & 74.6 & 71.5 & 73.0 & 85.1 & 79.3 & 82.1 \\
    \bottomrule
  \end{tabular}

\end{table}

\textbf{Effect of Arrow Agent.}
Removing the Arrow Agent leads to a severe performance degradation, with Hard Match F1 dropping by 39.8\% and Soft Match F1 decreasing by 24.8\%. This sharp decline highlights that arrow understanding provides indispensable structural constraints for graph construction. Without explicit arrow cues, the system loses reliable directionality and reaction boundary signals, resulting in widespread misassignment of reactant--product relationships. The pronounced drop in Hard Match further indicates that most failures stem from structurally invalid reaction grouping.

\textbf{Effect of Molecule-Recognition Agent.}
Excluding the Agent results in one of the most significant performance drops (Hard F1: 33.4\%, Soft F1: 59.8\%). This confirms that accurate molecular structure recognition is a prerequisite for chemistry-aware reasoning. Without reliable molecular identities, downstream modules cannot enforce stoichiometric consistency, reaction feasibility, or role assignment, causing errors that propagate globally across the reaction graph. Failures arise not only from incorrect reaction structure but also from fundamentally invalid chemical representations.

\textbf{Effect of Text-Recognition Agent.}
Removing the Agent reduces performance to 50.2\% Hard F1 and 71.3\% Soft F1. Textual annotations, including reagents, catalysts, and reaction conditions, provide critical semantic cues that complement visual perception. These cues help disambiguate molecular roles, distinguish parallel reactions, and refine reaction type inference. The relatively larger drop in Hard Match indicates that textual information primarily contributes to correct reaction-level grouping and assignment, rather than isolated entity recognition.

\textbf{Effect of Multigraph Fusion.}
Ablating the Multigraph-Fusion module leads to a substantial performance decrease (Hard F1: 50.4\%, Soft F1: 60.4\%), underscoring the necessity of jointly integrating heterogeneous evidence sources. Without fusion, spatial cues, chemical constraints, and VLM-derived semantic hypotheses remain fragmented, preventing the system from resolving conflicts and consolidating globally consistent reaction interpretations. In contrast, multigraph fusion provides a unified relational representation that enables coherent cross-modal evidence aggregation and reasoning. This result directly validates the central design choice of \textbf{MACReD}: modeling reaction parsing as a unified graph inference problem rather than a sequence of independent perceptual decisions.

\textbf{Effect of Post-Processing.}
Removing the Post-Processing stage causes a relatively modest performance drop (Hard F1: 73.0\%, Soft F1: 82.3\%), indicating that the core perception and reasoning modules already produce largely coherent reaction graphs. Nevertheless, post-processing plays a complementary role by enforcing chemical validity constraints, removing spurious or empty reactions, and standardizing outputs. This refinement step improves robustness and ensures chemically plausible final predictions, particularly in edge cases involving partial or noisy perceptual outputs.

Detailed ablation results across different layout types are further systematically provided in Appendix~\ref{ablation_details} (Table~\ref{tab:details_ablation_results}).
Additional qualitative reaction parsing examples, along with their corresponding visualizations, are also comprehensively presented in Appendix~\ref{sec:jsonandimage}.

\section{Conclusion}
\label{sec:conclusion}

We proposed \textbf{MACReD}, a hierarchical multi-agent framework for chemical reaction diagram parsing that tightly integrated multimodal perception with advanced chemistry-aware structured reasoning. By decomposing reaction understanding into specialized, collaborative agents and performing hierarchical graph evidence fusion, \textbf{MACReD} effectively overcame the limitations of VLMs in complex, structure-rich diagrams. Experiments on the RxnScribe benchmark consistently demonstrated outstanding SOTA performance and robustness across diverse reaction layouts, validating the effectiveness of multi-agent collaboration and explicit structural reasoning for scientific diagram understanding.
Subsequent work will focus on further improving coordination efficiency among agents, introducing more advanced chemistry-aware reasoning modules, and extending the framework to broader scientific domains.

\section{Limitations and Ethical Considerations}
\label{sec:ethical}
This study uses only publicly available chemical reaction datasets and does not involve human participants or any private data, so no informed consent or ethics approval is required. Potential applications include materials discovery and automated synthesis planning, where accurate multimodal perception and chemically informed structured reasoning are crucial. Current limitations include imperfect agent coordination, which may slightly affect overall performance on highly complex diagrams or challenging tasks.

\section{GenAI Discourse}

In the preparation of this manuscript and associated analyses, generative AI tools were employed solely for language polishing and formatting assistance. These tools helped improve clarity, conciseness, and readability, but did not contribute to the generation of experimental results, data analysis, or original scientific conclusions. All technical content, analyses, and interpretations were independently developed and verified by the authors.

The use of generative AI was carefully monitored to ensure accuracy and adherence to the scientific content. No proprietary or confidential data were exposed to AI tools, and all outputs were critically reviewed and edited to maintain factual correctness.

This disclosure aligns with transparent research practices and provides clarity on the role of AI-assisted writing in the manuscript preparation process.

\bibliographystyle{ACM-Reference-Format}

\bibliography{main}

\appendix

\section{Details of the Ablation Study}
\label{ablation_details}

\begin{table*}[htbp]
\centering
\caption{Detailed results of the ablation study for different layouts.}
\label{tab:details_ablation_results}
\resizebox{\textwidth}{!}{
\begin{tabular}{lcccccccccccc}
\toprule
\cmidrule(lr){2-13}
 & \multicolumn{3}{c}{Single Line} 
 & \multicolumn{3}{c}{Multiple Line} 
 & \multicolumn{3}{c}{Tree} 
 & \multicolumn{3}{c}{Graph} \\
\cmidrule(lr){2-4} \cmidrule(lr){5-7} \cmidrule(lr){8-10} \cmidrule(lr){11-13}
\textbf{Setting} 
 & Prec. & Rec. & F1 
 & Prec. & Rec. & F1
 & Prec. & Rec. & F1
 & Prec. & Rec. & F1 \\
\midrule
\multicolumn{13}{c}{\textbf{Hard Match}} \\
\midrule
\textbf{MACReD} 
& 85.2 & 81.5 & 83.3 
& 77.9 & 76.1 & 77.0 
& 78.1 & 78.0 & 77.0 
& 59.8 & 55.7 & 57.7 \\
\midrule
w/o Arrow & 40.8 & 45.7 & 43.1 & 48.2 & 43.5 & 45.7 & 40.0 & 30.0 & 34.3 & 11.1 & 6.6 & 8.2 \\
w/o Molecule-Recognition & 44.2 & 41.3 & 42.7 & 24.4 & 22.8 & 23.6 & 35.2 & 30.4 & 32.6 & 30.8 & 26.2 & 28.3 \\
w/o Text-Recognition & 58.0 & 55.4 & 56.7 & 44.9 & 43.5 & 44.2 & 57.2 & 51.6 & 54.3 & 19.3 & 18.0 & 18.6 \\
w/o Multigraph-Fusion & 59.3 & 56.1 & 57.7 & 44.7 & 43.7 & 44.2 & 57.9 & 51.3 & 54.4 & 20.1 & 17.0 & 18.4 \\
w/o Post-Processing & 84.9 & 80.4 & 82.6 & 75.1 & 73.3 & 74.2 & 77.4 & 68.7 & 72.8 & 57.1 & 49.2 & 52.9 \\
\midrule
\multicolumn{13}{c}{\textbf{Soft Match}} \\
\midrule
\textbf{MACReD} 
& 89.9 & 87.9 & 88.9 
& 87.9 & 86.3 & 87.1 
& 85.6 & 82.1 & 83.8 
& 81.4 & 73.7 & 77.4 \\
\midrule
w/o Arrow & 63.1 & 70.7 & 66.7 & 80.7 & 72.8 & 76.6 & 66.1 & 49.7 & 56.7 & 30.6 & 18.0 & 22.7 \\
w/o Molecule-Recognition & 72.1 & 67.4 & 69.7 & 51.1 & 47.8 & 49.4 & 62.5 & 49.0 & 54.9 & 57.7 & 49.2 & 53.1 \\
w/o Text-Recognition & 83.0 & 79.3 & 81.1 & 73.0 & 70.7 & 71.8 & 76.8 & 69.3 & 72.9 & 52.6 & 49.2 & 50.8 \\
w/o Multigraph-Fusion & 64.0 & 69.9 & 66.8 & 81.5 & 71.4 & 76.1 & 67.1 & 48.9 & 56.6 & 32.0 & 17.7 & 22.8 \\
w/o Post-Processing & 88.1 & 85.9 & 87.0 & 86.9 & 85.7 & 86.3 & 85.3 & 82.1 & 83.7 & 79.6 & 63.9 & 70.9 \\
\bottomrule
\end{tabular}
}
\end{table*}

Table~\ref{tab:details_ablation_results} presents a detailed ablation study of MACReD across four common layouts: Single Line, Multiple Line, Tree, and Graph. Evaluation metrics include Precision, Recall, and F1-score under both Hard Match and Soft Match criteria. The ablation results further highlight the importance of each module: removing arrow information (w/o Arrow) causes a significant drop in F1, particularly in the Graph layout, where it falls to 8.2\%, indicating that arrow information is critical for chemical reaction parsing. Similarly, removing the molecule recognition or text recognition modules substantially reduces performance, demonstrating that these components are core to reaction diagram understanding. Multigraph fusion (w/o Multigraph-Fusion) and post-processing (w/o Post-Processing) provide additional gains for complex structures, with post-processing showing the most notable improvements for Graph and Multiple Line layouts.
The performance suggests that the modules collectively enable multidimensional information fusion, effectively integrating perception and reasoning. This allows MACReD to handle visual and structural complexity as well as limited generalization, significantly enhancing the accuracy of reaction assembly.

\section{Datasets and Baselines}
\label{sec:dataandbaseline}

\subsection{Datasets and Preprocessing.}
We evaluate the proposed \textbf{MACReD} framework on the RxnScribe benchmark dataset, which is currently one of the most comprehensive publicly available resources for reaction diagram understanding. 
The dataset consists of 662 peer-reviewed chemistry articles collected from four representative journals: \textit{Journal of the American Chemical Society}, \textit{Journal of Organic Chemistry}, \textit{Organic Letters}, and \textit{Organic Process Research \& Development}. 
All reaction figures were automatically extracted from PDF files using the \texttt{pdffigures2} tool.
After preprocessing and filtering, the dataset contains reaction diagrams covering four representative layout styles: single-line, multi-line, tree-structured, and graph-based formats, reflecting the diversity of real-world chemical literature. 
To facilitate more accurate structural parsing, we further re-annotated the dataset with fine-grained arrow annotations using oriented bounding boxes (OBB), which enables robust detection of complex arrow geometries.

\subsection{Baselines.}
We compare \textbf{MACReD} against two categories of baseline methods. 
The first category comprises general-purpose VLM, including both closed-source and open-source models.
Specifically, we evaluate GPT-4o~\cite{Hurst2024GPT4oSC}, Gemini-2.5-Pro~\cite{geminiteam2025geminifamilyhighlycapable}, and QwenVL-Max~\cite{Qwen3-VL} as representative closed-source VLMs, as well as Qwen2.5-VL-7B~\cite{bai2025qwen25vltechnicalreport}, InternVL3-8B~\cite{zhu2025internvl3exploringadvancedtraining}, and MiMo-VL-7B~\cite{xiaomi2025mimo} as open-source alternatives. We additionally include ChemVLM-8B~\cite{li2025chemvlm} and Intern-S1-mini~\cite{bai2025interns1scientificmultimodalfoundation}, a chemistry-specialized VLM designed for molecular-level multimodal understanding.
All VLMs are directly prompted to perform reaction diagram interpretation in an end-to-end manner, without task-specific architectural modifications, explicit graph reasoning, or chemistry-aware constraints.

The second category consists of task-specific systems for reaction diagram parsing, including rule-based frameworks such as ReactionDataExtractor~\cite{wilary2021reactiondataextractor} and ReactionDataExtractor2~\cite{wilary2023reactiondataextractor}, which rely on handcrafted spatial heuristics and deterministic matching with limited learning-based detection, as well as learning-based pipelines like OChemR~\cite{M.Martori2022}, which uses a DETR-style vision transformer to detect molecules, text, and arrows before converting them into structured representations. More recent end-to-end approaches, including RxnScribe~\cite{qian2023rxnscribe} and RXNIM~\cite{chen2025towards}, leverage deep learning or multimodal large language models to directly translate reaction diagrams into machine-readable reaction representations.

For a fair and rigorous comparison, all baseline models are either retrained from scratch or fine-tuned using the same RxnScribe dataset, strictly adhering to identical training, validation, and test splits. This ensures that differences in performance reflect model capabilities rather than discrepancies in data exposure. In addition, we perform comprehensive ablation studies to systematically quantify the individual contributions of each key system component, including model architecture, feature representations, and reasoning modules. These analyses provide deeper insights into which elements are most critical for accurate reaction prediction and highlight potential avenues for further optimization.

\subsection{Implementation Details}
All experiments are conducted on a workstation equipped with an Intel(R) Xeon(R) Platinum 8352V CPU @ 2.10GHz, 251 GB of RAM, and an NVIDIA A40 GPU.
Our entire framework is implemented based on Gemini-2.5-Pro, a VLM chosen for its robust multimodal reasoning and strong instruction-following capabilities in scientific visual understanding tasks. During inference, we adopt a temperature of 0.7 and a nucleus sampling parameter top\_p=0.8 to balance generation diversity and factual consistency. The maximum output length is set to 8192 tokens, allowing sufficient capacity for structured, multi-turn reasoning and full reaction reconstruction.

\section{Prompt Design}
\label{sec:promptdesign}

This section presents the primary prompt templates used in our experiments (see Figure ~\ref{fig:PlannerAgent}, \ref{fig:MoleculeRecognitionrAgent}, and \ref{fig:ReactionCombinerAgent}). To ensure reproducibility, we provide the instructions for each specialized agent in the chemical reasoning pipeline. These prompts are categorized into three types:

Planner Agent Prompts: These instructions guide the model to analyze the user's chemical image and question, and determine which expert modules should be activated for the task. They enforce strict JSON output without additional explanation.

Molecule-Recognition Agent Prompts: These instructions explicitly direct the model to verify and correct the SMILES representations of detected molecules, ensuring strict chemical validity in terms of bond connectivity, stereochemistry, and other fundamental rules. The output is a structured JSON array containing molecular labels, bounding boxes, and canonical SMILES strings.

Reaction-Combine Agent Prompts: These instructions explicitly instruct the model to refine and prune the chemical reaction network graph, constructing fully chemically valid reactions from detected molecules, arrows, text, and identifiers. They enforce strict adherence to chemical rules, outputting a complete and consistent JSON array of reactants, products, conditions, and arrows.

Together, these prompts provide a structured framework for stepwise chemical reasoning, combining perception, semantic grounding, and reaction assembly in a reproducible and verifiable manner.

\begin{figure*}[!ht] 
    \begin{tcblisting}{
        title=Planner Agent Prompt,
        colback=white,
        colframe=black!75,
        listing only,
        width=\textwidth,
        listing options={
            basicstyle=\small\rmfamily,
            breaklines=true,
            columns=fullflexible,       
            keepspaces=true
        }
    }
You are a Task Planning Agent specialized in chemical reasoning. The user will provide a chemical-related image along with a specific question. 

Your goal is to analyze both the user's question and the content of the image, and determine which of the following expert modules need to be activated for this task:
- molecule_expert: responsible for molecular detection and structural recognition.
- arrow_expert: responsible for detecting reaction arrows and analyzing their directionality.
- text_expert: responsible for detecting textual elements and performing OCR recognition.
- reaction_expert: responsible for assembling chemical reaction pathways based on detected molecules, arrows, and text.

For example:
- If the task requires full chemical reaction recognition, return:
  {"molecule_expert": true, "arrow_expert": true, "text_expert": true, "reaction_expert": true}
- If the task only requires converting a molecular diagram to a SMILES string, return:
  {"molecule_expert": true, "arrow_expert": false, "text_expert": false, "reaction_expert": false}

Output the result strictly in valid JSON format, following the structure:
{"plan": {"molecule_expert": true, "arrow_expert": false, "text_expert": true, "reaction_expert": true}}

Do not include any explanations or additional text---only output the JSON.

    \end{tcblisting}
\caption{Planner Agent prompt for context-aware agent selection.}
\label{fig:PlannerAgent}
\end{figure*}

\begin{figure*}[!ht] 
    \begin{tcblisting}{
        title=Molecule-Recognition Agent Prompt,
        colback=white,
        colframe=black!75,
        listing only,
        width=\textwidth,
        listing options={
            basicstyle=\small\rmfamily,
            breaklines=true,
            columns=fullflexible,       
            keepspaces=true
        }
    }
You are an expert in chemical molecular structure recognition. 

Based on the chemical image provided by the user, the positions of molecular structures detected by the molecule_expert, and the SMILES strings output by the Image2SMILES module, verify and correct the SMILES representation of each chemical molecule in the image. Pay particular attention to potential chemical inconsistencies, such as bond connectivity, stereochemistry, and other basic chemical rules. Additionally, output the name and structural representation of each molecule.

Finally, return a valid JSON array where each element contains the following three fields:
- "label": fixed as "molecule"
- "bbox": an array [x_min, y_min, x_max, y_max]
- "smiles": the SMILES string of the molecule

Return JSON only. Do not include any explanations.
    \end{tcblisting}
\caption{Molecule-Recognition Agent prompt for recognition molecular entities.}
\label{fig:MoleculeRecognitionrAgent}
\end{figure*}

\begin{figure*}[!ht]
    \begin{tcblisting}{
        title=Reaction-Combiner Agent Prompt,
        colback=white,
        colframe=black!75,
        listing only,
        width=\textwidth,
        listing options={
            basicstyle=\small\rmfamily,
            breaklines=true,
            columns=fullflexible,       
            keepspaces=true
        }
    }
Below is the chemical reaction network graph generated by the agent, including all nodes (molecule / text / identifier / arrow) and the types and weights of their edges. The edge type mapping is as follows:
    - REL_REACTANT_TO_COND = 0  - REL_COND_TO_PRODUCT = 1   - REL_REACTANT_TO_PRODUCT = 2 
    - REL_NO_EDGE = 3   - NUM_RELATIONS = 4   - REL_REACTANT_TO_ARROW = 5   - REL_ARROW_TO_PRODUCT = 6

Your task is to prune and refine this graph based on edge weights and chemical properties to construct chemically valid and complete reactions. The final reactions must adhere to chemical rules, even if some information is missing or redundant due to prior segmentation of complex graphs.

Steps to follow:
- Typically, the number of reactions should not be less than the number of arrows. If there are two arrows with no intermediate objects, treat them as a single arrow and attempt to complete the reaction. If it is impossible to find corresponding reactants or products, discard that arrow.
- Preliminary results may contain inconsistencies. Use chemical knowledge to check each reaction and fill in missing components.
- Ensure that reaction conditions do not include unrelated information.
- If an identifier is associated with a molecule, remove the identifier from the reaction and replace it with the corresponding molecule to avoid duplicate representation.

Requirements:
- Each reaction must include the fields: reactants, products, conditions, arrow.
- The reactants and products fields must not be empty; conditions and arrow may be empty.
- Values in reactants and products can be molecule, identifier, or text.
- The arrow field can only contain arrow elements.

Return strictly a complete, valid JSON array, containing only the final results. Do not include any additional explanation or data. Example format:
[
    {
        "reactants": [{"label": "molecule", "bbox": [38, 2, 434, 234]}],
        "products": [{"label": "molecule", "bbox": [912, 14, 1309, 231]}],
        "conditions": [
            {"label": "text", "bbox": [515, 66, 855, 126]},
            {"label": "text", "bbox": [577, 172, 780, 223]}
        ],
        "arrow": [{"label": "arrow", "bbox": [513, 155, 880, 153, 880, 130, 513, 132]}]
    },
    {
        "reactants": [
            {"label": "text", "bbox": [246, 48, 370, 99]},
            {"label": "identifier", "bbox": [482, 48, 540, 96]}
        ],
        "products": [{"label": "molecule", "bbox": [837, 9, 1095, 132]}],
        "conditions": [
            {"label": "text", "bbox": [597, 3, 759, 50]},
            {"label": "text", "bbox": [592, 87, 767, 137]}
        ],
        "arrow": [{"label": "arrow", "bbox": [585, 76, 789, 75, 789, 57, 585, 59]}]
    }
]
    \end{tcblisting}
\caption{Reaction-Combiner Agent prompt for refining reaction graphs into chemically valid reactions.}
\label{fig:ReactionCombinerAgent}
\end{figure*}

\section{Results Visualization}
\label{sec:jsonandimage}

\begin{figure*}[!ht]
\centering
\includegraphics[width=0.9\textwidth]{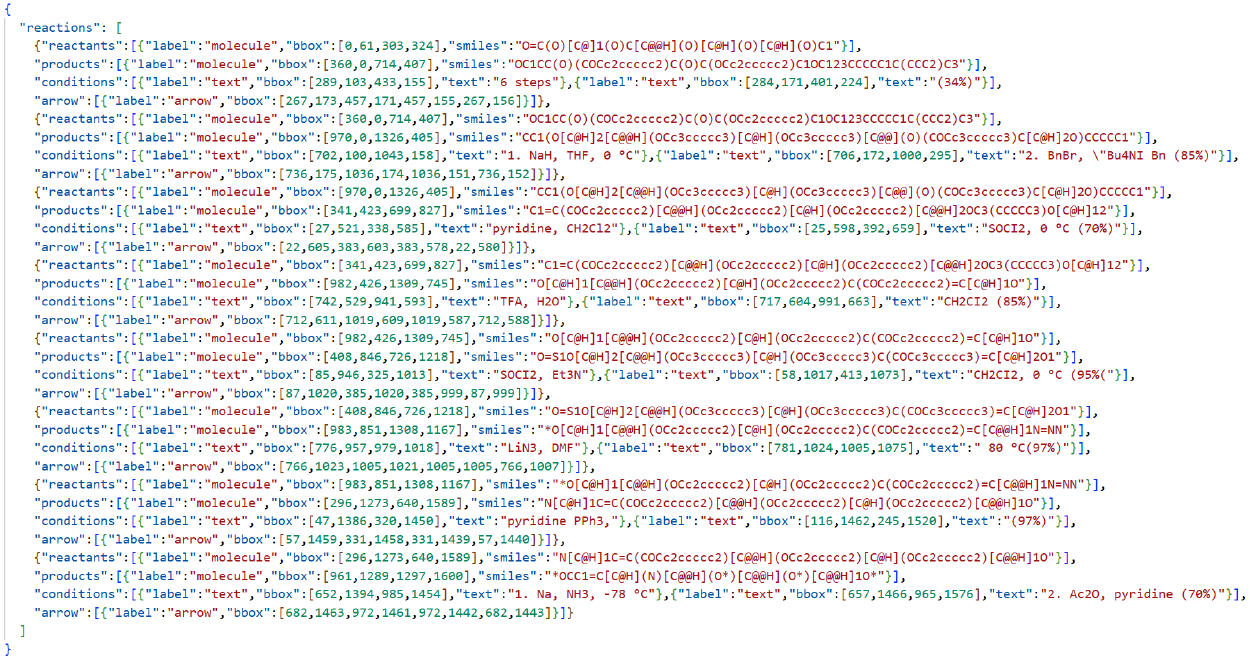}
\caption{Structured JSON output from Reaction Diagram Parsing, showing identified reactants, products, conditions, and arrows in machine-readable format.}
\label{fig:resultjson}
\end{figure*}

\begin{figure*}[!ht]
\centering
\includegraphics[width=0.9\textwidth]{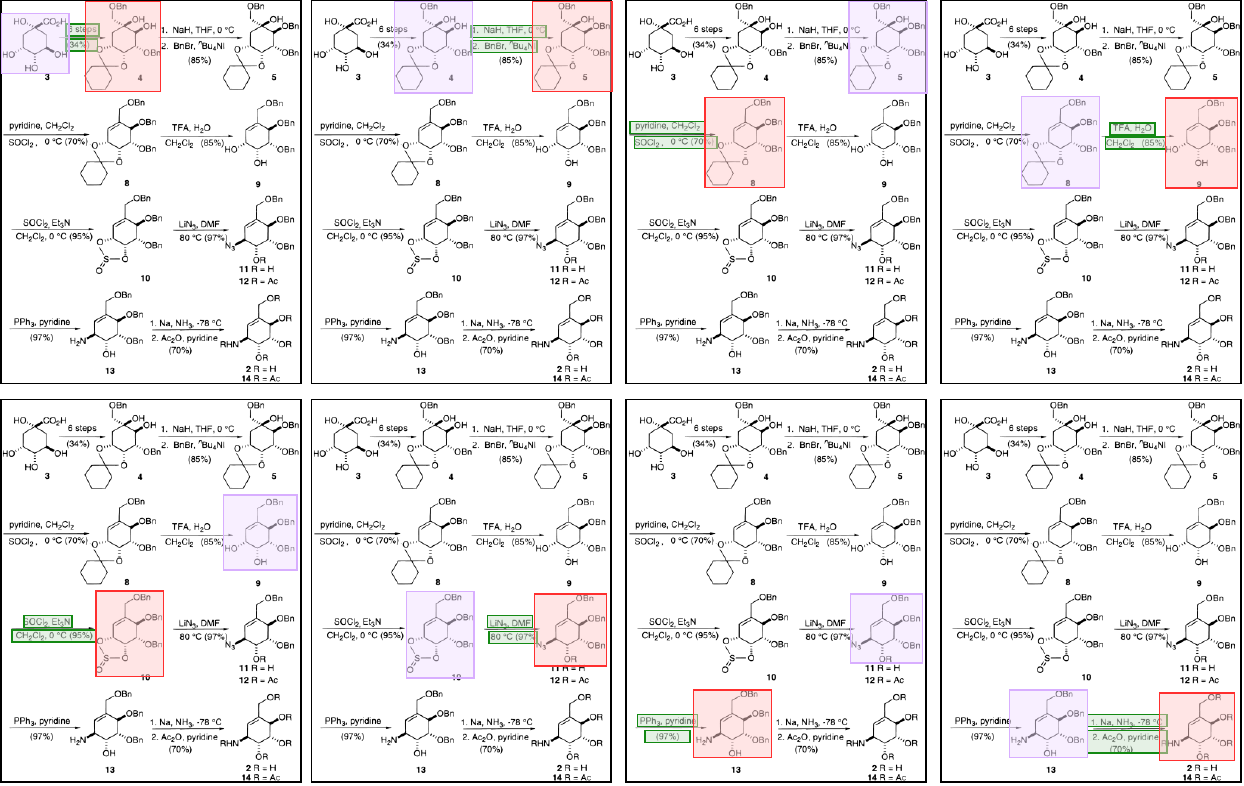}
\caption{Visualization of Reaction Diagram Parsing JSON output, illustrating the spatial layout of molecules, conditions, and reaction arrows.}
\label{fig:resultima}
\end{figure*}

In this section, we present representative visualizations of the Reaction Diagram Parsing outputs. Figure~\ref{fig:resultjson} shows the structured JSON format generated by our system, detailing the detected reactants, products, reaction conditions, and arrows in a machine-readable form. Figure~\ref{fig:resultima} illustrates the spatial layout of these elements, providing an intuitive view of how molecules, conditions, and reaction arrows are positioned within the original diagram.

\begin{figure*}[!ht]
\centering
\includegraphics[width=0.9\textwidth]{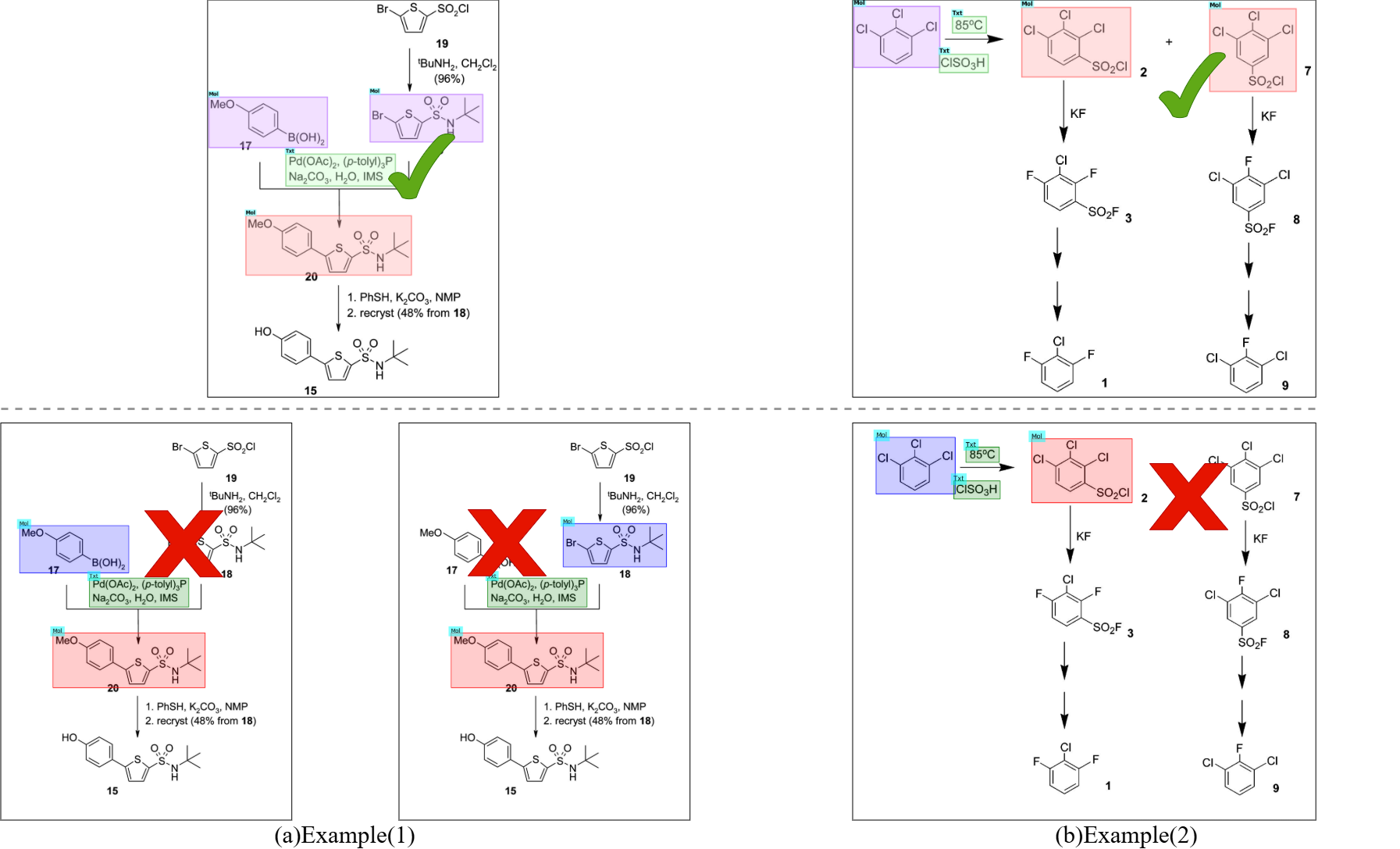}
\caption{Qualitative comparison between \textbf{MACReD} (top) and \textbf{RxnScribe} (bottom). 
\textbf{MACReD} produces more complete reaction predictions.}
\label{fig:compare}
\end{figure*}

Additionally, Figure~\ref{fig:compare} provides a qualitative comparison between our system, \textbf{MACReD}, and RxnScribe. The comparison highlights that \textbf{MACReD} generates more complete and chemically consistent reaction predictions, demonstrating the effectiveness of our multi-agent reasoning and structured output approach.

\end{document}